\documentclass[conference]{IEEEtran}
\usepackage{amsmath,amssymb,amsfonts}
\usepackage{amsthm}
\usepackage{graphicx}
\usepackage{cite}

\newtheorem{proposition}{Proposition}
\theoremstyle{remark}

\usepackage{xcolor}
\usepackage{booktabs}
\usepackage{algorithm}
\usepackage[noend]{algpseudocode}
\newcommand{\NUM}[1]{\ifmmode\text{\textbf{\textcolor{red}{[#1]}}}\else\textbf{\textcolor{red}{[#1]}}\fi}
\usepackage[normalem]{ulem}
\newif\ifreview \reviewfalse

\usepackage{multirow}
\usepackage{graphicx}
\usepackage{subcaption} 
\usepackage{caption}
\IEEEoverridecommandlockouts
\begin{document}

\title{Haptic Probing for Subsurface-Aware\\
Force Distribution on Layered Soft Tissue}

\author{
\IEEEauthorblockN{
Lei Shi$^{*}$,
Xiaoyu Wang$^{*}$,
Cheng Zhou$^{*\dagger}$,
Yongjian Liu,
Jing Wang,\\
Dongsheng Zhang,
Yonghui Zhu,
Xiong Li$^{\dagger}$%
}
\thanks{$^{*}$These authors contributed equally to this work, $^{\dagger}$Corresponding author.}
}

\maketitle

\begin{abstract}
Force-regulated contact with layered soft tissue fails most often where the
subsurface composition changes: controllers infer the substrate from the loading
signal alone, so the estimate guiding them lags the change---more acutely, we
argue, over compliant than over bone-backed tissue. We remove that lag with a
separate \emph{haptic} channel. Under a light preload, a brief low-amplitude
excitation lets the local dynamic admittance discriminate muscle from bone-backed
tissue relative to a per-subject baseline, before task force is applied---active
haptic sensing used to commit a control decision rather than to characterise
tissue. A force- and tactile-driven distribution law then consumes that regime
throughout execution:
an instantaneous allocation equalises pressure across the engaged fingertips,
while the regime sets the admissible pressure and hence the area the hand must
recruit to deliver the target force---making patch area a controlled variable
rather than a by-product. Probing, switches and recruitment are priced in a
partitioned passivity budget, so safety monitoring cannot suppress the probe and a
refused recruitment forfeits force tracking rather than the pressure bound. We
report force tracking and pressure uniformity over curved, migrating contact on
three human subjects, with regime labels from anatomical landmarks.
\end{abstract}


\section{Introduction}\label{sec:intro}

Robots that assist with physiotherapy and massage~\cite{harada2025massage},
ultrasound acquisition~\cite{jiang2021forceprobe}, or surgical retraction must
hold a regulated force against soft tissue for extended intervals. Force control
for these tasks is well developed, but against a \emph{single} contact regime: a
target force on one quasi-homogeneous surface, with whatever lies beneath treated
as a disturbance. The substrate is instead a layered viscoelastic stack---skin
over fat over muscle, in places muscle over bone---and as the tool travels the
contact is a patch whose area changes with curvature while the tissue stores and
returns energy. Changes in subsurface support can cause the measured contact force to deviate from its reference and reduce conformance as the tool moves across the tissue.

The dominant controllers share one blind spot. Impedance and admittance control
with their adaptive extensions~\cite{hogan1985impedance, li2014impedance,
abudakka2020variable}, switched and hybrid schemes triggered on measured force,
velocity or gait phase~\cite{ott2014hybrid, zotovic2005impact, cheng2018switched,
warner2016switched, 10160817}, and tactile servoing of surface
geometry~\cite{lepora2022digitac, lloyd2023pose, fan2023tacvgnn, wang2024mpc} all
infer the substrate from the loading signal alone, so the right behaviour arrives
only after loading begins (Section~\ref{sec:related})---yet the subsurface class
is what decides whether to yield into compliant tissue or cap the force and
spread the load.

That lag is not uniform across materials, which is the crux of our argument. An
online estimate must accumulate contact-signal history before converging; a few
samples suffice in stiff, lightly damped media, but soft tissue is heavily damped
and viscoelastic~\cite{wheatley2016muscle}, so it settles slowly. Our elements
then follow from one another rather than stacking: the soft regime makes the lag
acute, which makes a feedforward commitment worthwhile, which requires a channel
reporting before task force, which must be read relative to each subject because
absolute soft-tissue mechanics vary too widely to threshold.

We therefore resolve the regime first, by haptic means: under a light preload, a
brief low-amplitude vibration excites the local dynamics, and the resulting
admittance separates bone-backed from muscle tissue \emph{relative} to a
same-subject baseline, which absorbs inter-subject variation into the reference
rather than the decision boundary. Both channels the method uses are haptic---an
actively injected excitation read at the force port, and per-finger contact area
read from the tactile arrays---and both are used to \emph{act} rather than to
report. The probe is far smaller than the task force and completes before it, so
the ordering is probe-before-task-loading rather than
probe-before-any-contact---it runs in light contact, and what makes the prior
feedforward is that it is committed before the force it governs is commanded. And
what the discrete class commits is not a gain set but a \emph{distribution
objective}: how much pressure the tissue beneath the patch may carry, hence how
much area the hand must recruit. Because a passivity- or barrier-based safety layer~\cite{shi2026escape, shi2025fair}
cannot distinguish an injected probe from a disturbance, we reserve a bounded, separately
accounted energy band for it.

A related capability exists in hard-tissue drilling, where vibration detects the
cortical-to-cancellous transition~\cite{lee2004bonedrilling, dai2021vibration},
but the physics differ: there a high-energy cutting by-product crosses layers
differing by orders of magnitude in modulus (GPa for cortical
bone~\cite{choi1990bonemoduli}), giving a sharp signature detected reactively.
Ours is low-amplitude excitation on uncut kPa-range tissue, heavily damped, so the
response is weak, broad, and state-dependent rather than a material
fingerprint~\cite{wheatley2016muscle}---which is why we read it against a
per-subject baseline.

This paper makes the following contributions.
\begin{itemize}
\item \textbf{Haptic probing.} An active haptic probe---an injected excitation
read as a local dynamic admittance---resolving the subsurface regime \emph{before
task loading}, discriminated \emph{relative to a per-subject baseline} rather than
by absolute stiffness or cross-subject classification. What separates it from
haptic tissue characterisation is purpose: the reading is binary, taken inside the
contact task, and exists to be consumed by a controller rather than reported
(Section~\ref{sec:probe}).
\item \textbf{Force control on the probe's output.} A force-distribution law in
which that regime indexes how the resultant is spread over the patch: an
instantaneous allocation equalises pressure across the engaged fingertips, while
the regime sets the admissible pressure and hence, through a feasibility
condition, the area the hand must recruit. Capping the force and spreading it is
thereby derived rather than asserted, and patch area becomes a controlled variable
closed through distributed haptic feedback from the tactile arrays
(Section~\ref{sec:control}).
\end{itemize}


\section{Related Work}\label{sec:related}

\subsection{Impedance, variable, and switched force control}
Impedance control~\cite{hogan1985impedance} and its adaptive, variable, and
learned variants estimate the environment on-line and reshape the force--motion
relationship accordingly~\cite{li2014impedance, abudakka2020variable, buchli2011learning},
including whole-body impedance coordination over uncertain terrain~\cite{11266909}.
Switched and hybrid impedance methods make the change discrete rather than
continuous: \cite{ott2014hybrid} unifies impedance and admittance causality with a
passivity guarantee; \cite{zotovic2005impact} switches impedance parameters at
impact; \cite{cheng2018switched} toggles on a binary contact-force condition for
beating-heart surgery; \cite{warner2016switched} indexes gait phases of a
prosthesis through a finite-state machine. Learned or offline-scheduled stiffness
profiles, and phase-indexed schemes such as the last, do carry a genuinely
feedforward gain---the schedule exists before contact---so the distinction is not
feedforward versus feedback but what the index is derived from. Theirs is
\emph{task-derived}: time or phase along a cycle the robot generates and can
anticipate, which cannot re-index when the substrate departs from what it was
tuned on. Ours is \emph{environment-derived}, a measurement at a locus of an
aperiodic structure the robot neither controls nor predicts, which must therefore
be sensed before that locus is loaded. We borrow the passivity of the underlying
controller from~\cite{ott2014hybrid} and the dissipativity of a parameter switch
from~\cite{zotovic2005impact}, and treat energy \emph{boundedness} at the switch
instant with~\cite{liberzon2003switching} and the accounting of
Section~\ref{sec:passivity}---boundedness, not asymptotic stability.

\subsection{Tactile servoing and haptic sensing}
Tactile servoing regulates the geometry of surface contact---pose, shear, and
slip---using high-resolution optical or capacitive
sensors~\cite{lepora2022digitac, lloyd2023pose, fan2023tacvgnn, wang2024mpc}.
Where these works speak of ``depth'' they mean indentation of the sensor skin, and
by ``soft'' they mean that skin, not the substrate. The distinction is not between
sensors but between index signals: tactile servoing regulates contact pose and
shear indexed by the sensor's own surface reading, whereas we regulate the
\emph{distribution} of force across the patch indexed by the tissue class beneath
it---precisely what a surface signal does not carry, and why the allocation of
Section~\ref{sec:control} cannot be recovered by servoing harder on surface
features.

\subsection{Deformable manipulation and subsurface sensing}
Force control on viscoelastic bodies with area contact is established.
CATCH-FORM~\cite{ma2025catchform3d} regulates contact
force and deformation on a 3D viscoelastic object with a single compliance-aware
model of the contacted body; \cite{tiezzi2007viscoelastic} models viscoelastic
soft-fingertip contact with a time-evolving contact area. Both regulate a
resultant, and its deformation, against one model of the engaged body held fixed
for the contact. Here the substrate is heterogeneous in depth, so the applicable
model changes \emph{within} a traverse, and the regulated quantity is not only the
resultant but its distribution.

Distributing a resultant over several contacts is itself classical: grasp analysis
optimises contact forces under equilibrium and friction
constraints~\cite{siciliano2016robotics} with standard
solvers~\cite{boyd2007fast}, and~\eqref{eq:alloc} shares that structure
and claims nothing new in it. What differs is the index. Grasp force optimisation
is indexed by the grasp's own geometry and friction, both measurable at the
contact; ours is indexed by the tissue class beneath the patch, which no contact
measurement reveals and which changes along the traverse, so the pressure bound is
a per-locus consequence of what was probed rather than a fixed property of the
mechanism.

Haptic subsurface sensing is the closest counter-family, and the one our probe
belongs to. Robotic palpation maps stiffness by rolling or indenting to localise
hard inclusions~\cite{nichols2015segment}; active exploration chooses where to
probe next or estimates a stiffness field during sustained
contact~\cite{ayvali2017utility}; active-probing perception modulates a probe to
improve haptic discrimination~\cite{sornkarn2017softprobe}; shear-wave
elastography recovers tissue mechanics without
contact~\cite{sarvazyan1998shearwave}. These share our sensing principle but not
its role: they sense through sustained or quasi-static contact and consume the
result for localisation, imaging or perception---the haptic reading is the
\emph{output}. Prior soft-tissue vibration work targets friction reduction during
insertion, not backing discrimination~\cite{khalaji2013needle}. Ours is the same
class of measurement put in the loop instead: a low-energy probe completed
\emph{before task loading}, read \emph{relative} to a per-subject baseline, and
wired \emph{feedforward} into the distribution law, where it sets the pressure
bound and hence the recruited area. The haptic reading is an \emph{index}, not a
report.

\section{Method}\label{sec:method}

\subsection{Problem formulation}\label{sec:problem}
A multi-fingered hand on a force-controlled arm traverses layered viscoelastic
tissue while regulating a target \emph{resultant} normal force $f_d>0$. Finger
$i$ contacts over a patch of area $a_i(t)$, measured by its tactile array, at
normal position $x_i$ and force $f_{e,i}\ge0$, positive in compression. Each
finger is one lumped normal port; with engaged set $S$,
\begin{equation}
f_e=\sum_{i\in S}f_{e,i},\qquad A(t)=\sum_{i\in S}a_i(t),\qquad
p_i=\frac{f_{e,i}}{a_i}.
\label{eq:ports}
\end{equation}
Within a port the tissue is distributed and viscoelastic, so $x_i$, $f_{e,i}$ are
that finger's normal response, not a spatial field, and $p_i$ is the patch
average rather than a peak taxel value. We call the joint
objective of regulating $f_e$ toward $f_d$ \emph{subject to a regime-selected
compliance}, while holding the pressure distribution $\{p_i\}$ within a
regime-selected bound over the migrating patches (Section~\ref{sec:control}),
\emph{force wrapping}. It is a two-criterion objective and we score it as one:
the tracking error $|f_e-f_d|$ and the uniformity $U=\max_i p_i/\bar{p}$, with
$\bar{p}=f_e/A$ and $U\!\ge\!1$ by construction. Neither alone suffices---a resultant can be
tracked while one fingertip carries most of it---which is what makes the
distribution, and not only the resultant, the controlled quantity. We assume throughout: \textbf{A1} (quasi-static area): $a_i$ varies slowly
relative to the allocation update, so~\eqref{eq:alloc} treats it as a parameter
within a sample. \textbf{A2} (local linearisation): near the operating point,
the tissue under port $i$ has effective stiffness $k_{t,i}$ about an unloaded
position $x_{s,i}$ (index dropped for a single port). \textbf{A3} (state
continuity): $x_i$ and $\dot{x}_i$ are continuous at switches and recruitment,
which change references rather than apply impulses. \textbf{A4} (patch-average
pressure): $p_i$ is averaged over port $i$'s patch; peak taxel pressure is not
regulated. \textbf{A5} (quiescent standoff): an unengaged port rests at its
standoff reference ($\dot{x}_j=0$, $x_j=x_{d,j}$), so its energy
in~\eqref{eq:V} is zero, and its impedance switches, acting on a zero state,
contribute nothing to~\eqref{eq:dV}; a non-quiescent port must be counted, and
Section~\ref{sec:experiments} verifies quiescence. \textbf{A6} (dissipative
target): $M_d^{r},D_d^{r},K_d^{r}>0$ for all $r$, so the target dynamics
dissipate within each regime. All $n$ ports always run~\eqref{eq:imp},
unengaged ones at standoff with $f_{d,j}=f_{e,j}=0$; $S$ enters
only~\eqref{eq:alloc}, so recruitment changes a reference, not the set over
which energy is accounted.

The task decomposes along two axes:
\begin{itemize}
\item a \emph{depth axis}---the subsurface regime $r\in\{\textsc{bone},
\textsc{muscle}\}$, sensed by the haptic probe before task loading and used
feedforward;
\item a \emph{surface axis}---the contact-patch pose and force, tracked
continuously by the tactile arrays and used feedback.
\end{itemize}

\subsection{Haptic probing before task loading and relative discrimination}\label{sec:probe}
Immediately before task loading, the tool applies
a low-amplitude excitation $u_{\mathrm{p}}(t)$ over a short window and records
the response. Probing is not done in free space and it runs under a light preload $f_{\mathrm{pre}}$ with
$f_{\mathrm{pre}}+\varepsilon_{\mathrm{p}}\ll f_d$, and the bandwidth is capped
at $\omega_{\max}$. 
We characterize the tissue by its dynamic
admittance
\begin{equation}
Y(\omega)\;=\;\frac{\dot{X}(\omega)}{F_e(\omega)},
\end{equation}
a lumped single-DOF transfer function and the haptic quantity the probe actually
returns, mapped to a fixed-dimensional feature vector $\phi\in\mathbb{R}^{d}$: the
dominant response-peak frequency, an effective damping ratio, and the
band-averaged response magnitude.

Absolute $\phi$ varies strongly across subjects, so classification is
\emph{relative} to a per-subject in-situ baseline: at the start of a session the
tool records references $\phi^{\textsc{bone}}_0$, $\phi^{\textsc{muscle}}_0$ at
anatomically identified sites on that subject. The regime at a new site is the
nearer reference in a Mahalanobis-type metric $d(\cdot,\cdot)$,
\begin{equation}
r \;=\; \arg\min_{c\,\in\,\{\textsc{bone},\textsc{muscle}\}}\;
d\!\left(\phi,\;\phi^{c}_0\right),
\end{equation}
with the decision reported together with the confidence $\lvert m\rvert$,
\begin{equation}
m \;:=\; d\!\left(\phi,\;\phi^{\sigma}_0\right)
       - d\!\left(\phi,\;\phi^{\sigma'}_0\right)
\label{eq:margin}
\end{equation}
is the \emph{signed} margin against the currently committed regime $\sigma$, so
that $m>0$ favours the alternative $\sigma'$. The sign is load-bearing: the switching test of
Section~\ref{sec:control} is stated on $m$, not its magnitude.

\subsection{Feedforward switched control with regime-conditioned wrapping}\label{sec:control}
Each engaged port is regulated by a target-impedance law in the error coordinate
$e_i = x_i - x_{d,i}$, where $x_{d,i}$ is that port's reference normal position:
\begin{equation}
M_d^{r}\,\ddot{e}_i + D_d^{r}\,\dot{e}_i + K_d^{r}\,e_i \;=\;
\alpha f_{d,i} + \alpha_{\mathrm{p}} u_{\mathrm{p},i} - f_{e,i} ,
\label{eq:imp}
\end{equation}
where force reference $f_{d,i}$ is set by the allocator
of~\eqref{eq:alloc} below. The law realizes an impedance
causality whose passivity is the
property we borrow from~\cite{ott2014hybrid}. This matters for the spectrum below, because the largest stably rendered
stiffness $K_d^{\max}$ bounds how position-dominated the \textsc{bone} setting
can be through $\eta_r$ of~\eqref{eq:spectrum}. Its parameters and target are selected by the
probed regime,
\begin{equation}
\Theta_r \;=\; \{M_d^{r}, D_d^{r}, K_d^{r}, x_d^{r}, f_d^{r}\} \;=\; \Pi(r).
\end{equation}

The two regimes differ in where they sit on a spectrum between force-dominated
and position-dominated contact. Representing the engaged tissue near the
operating point by an effective stiffness $k_t$ about an unloaded surface
position $x_s$, the equilibrium of~\eqref{eq:imp} carries
\begin{equation}
f_e^{\ast} \;=\; \eta_r\, f_d^{\,r} \;+\; (1-\eta_r)\,k_t\bigl(x_d^{\,r}-x_s\bigr),
\quad
\eta_r \;=\; \frac{k_t}{k_t+K_d^{\,r}} .
\label{eq:spectrum}
\end{equation}
A small $K_d^{r}$ gives $\eta_r\!\to\!1$: force-dominated contact,
$f_e^{\ast}\!\to\!f_d$, the tool yielding into compliant tissue---the
\textsc{muscle} strategy. A large $K_d^{r}$ with a shallow $x_d^{r}$ gives
$\eta_r\!\to\!0$: penetration is bounded by the commanded depth, so force is
capped rather than driven to $f_d$---the \textsc{bone} strategy.

Given the
regime and the measured patch areas, the per-port references are the solution:
\begin{equation}
\begin{split}
\{f_{d,i}\}=\operatorname*{arg\,min}_{f}\ &\sum_{i\in S}\frac{f_i^{2}}{\underline{a}_i}\\[-2pt]
\text{s.t.}\ \ &\sum_{i\in S}f_i= f_d^{\,r},\quad
0\le f_i\le p^{\max}_r\,\underline{a}_i ,
\end{split}
\label{eq:alloc}
\end{equation}
where $\underline{a}_i=c\,a_i$, $c\in(0,1]$, is a conservative lower bound on the
measured area taken \emph{proportional} to it. Proportionality is what carries the
objective's guarantee from the surrogate to the measurable quantity, and it is
load-bearing rather than notational. With the resultant fixed, the unconstrained
minimiser is $f_i^{\ast}=f_d^{\,r}\underline{a}_i/\sum_j \underline{a}_j$, which
equalises $f_i/\underline{a}_i$; since $c$ cancels,
\begin{equation}
p_i=\frac{f_i^{\ast}}{a_i}
=\frac{f_d^{\,r}\,c\,a_i}{c\sum_j a_j\;a_i}
=\frac{f_d^{\,r}}{\sum_j a_j}
\label{eq:puniform}
\end{equation}
is likewise equal across ports, so the \emph{measured} pressure
of~\eqref{eq:ports} is uniform, not merely the surrogate. A fixed additive
margin would not do this: it equalises $f_i/\underline{a}_i$ while leaving $p_i$
spread by the ratio of the margins, so the guarantee would hold only for a
quantity we do not measure.
Beyond that, \eqref{eq:alloc} allocates the \emph{nominal} target: the safety
scaling enters once, in~\eqref{eq:imp}, so the two run in sequence rather than
circularly and the commanded $\alpha f_{d,i}$ only tightens a pressure
\eqref{eq:alloc} already bounds. Uniform pressure is thus the solution of
\eqref{eq:alloc} under a proportional bound, not a constraint bolted onto it.

The regime therefore enters through three channels, not one: the per-port
impedance $\Theta_r$, the pressure bound $p^{\max}_r$, and the engaged set
$S_r$. The last two are coupled, because \eqref{eq:alloc} is feasible only if
there is enough area to carry the resultant under the bound:
\begin{equation}
\sum_{i\in S}\underline{a}_i\ \ge\ A^{\,r}_{\mathrm{req}}
:=\frac{f_d^{\,r}}{p^{\max}_r}.
\label{eq:areq}
\end{equation}
which under the proportional bound reads $c\sum_{i\in S}a_i\ge
A^{\,r}_{\mathrm{req}}$, so measured area sets the requirement up to the known
$c$. When it fails the controller recruits a further finger, subject to the
admission rule of Section~\ref{sec:passivity}, until it holds, a hysteresis band
preventing chatter as the patches migrate. This is what makes the \textsc{bone}
strategy concrete: a tighter $p^{\max}_r$ raises $A^{\,r}_{\mathrm{req}}$ and
therefore recruits area, so ``cap the force and spread it'' is a recruitment law
derived from the pressure bound rather than a side effect of a stiff setting, and
$A(t)$ becomes a controlled variable closed through the tactile arrays. With the
fallback of Proposition~\ref{prop:failsafe}, \eqref{eq:alloc} is feasible at
every sample independently of the previous one---recruitment restores
\eqref{eq:areq}, or, if refused, the resultant is lowered until it holds with
equality---which is what makes feasibility robust to the shrinking $a_i$ that
patch migration produces.

Because $r$ is available before task loading, the controller starts from
$\Theta_r$ at the first loaded sample rather than converging to it from a
nominal $\Theta$.

A reading taken at one locus goes stale in space at a regime boundary,
reintroducing along the spatial axis the staleness we criticise in reactive
control. We therefore re-probe rather than commit once: the controller monitors
the signed margin $m$ of~\eqref{eq:margin} and the contact-signal residual, and
on detecting a boundary unloads to the probing preload, probes the new locus, and
only then commits $\Theta_r$ there. A switch is proposed only when $m$ exceeds a
hysteresis threshold $m_{\mathrm{on}}>0$ above the margin noise, so an exact
boundary cannot generate a switch sequence; on abstain with no prior commitment
we default to \textsc{bone}, the force-capping choice. Every $\Theta_r$ is thus
set before its segment is loaded.

\begin{figure}[htbp]
\centering
\includegraphics[width=\linewidth]{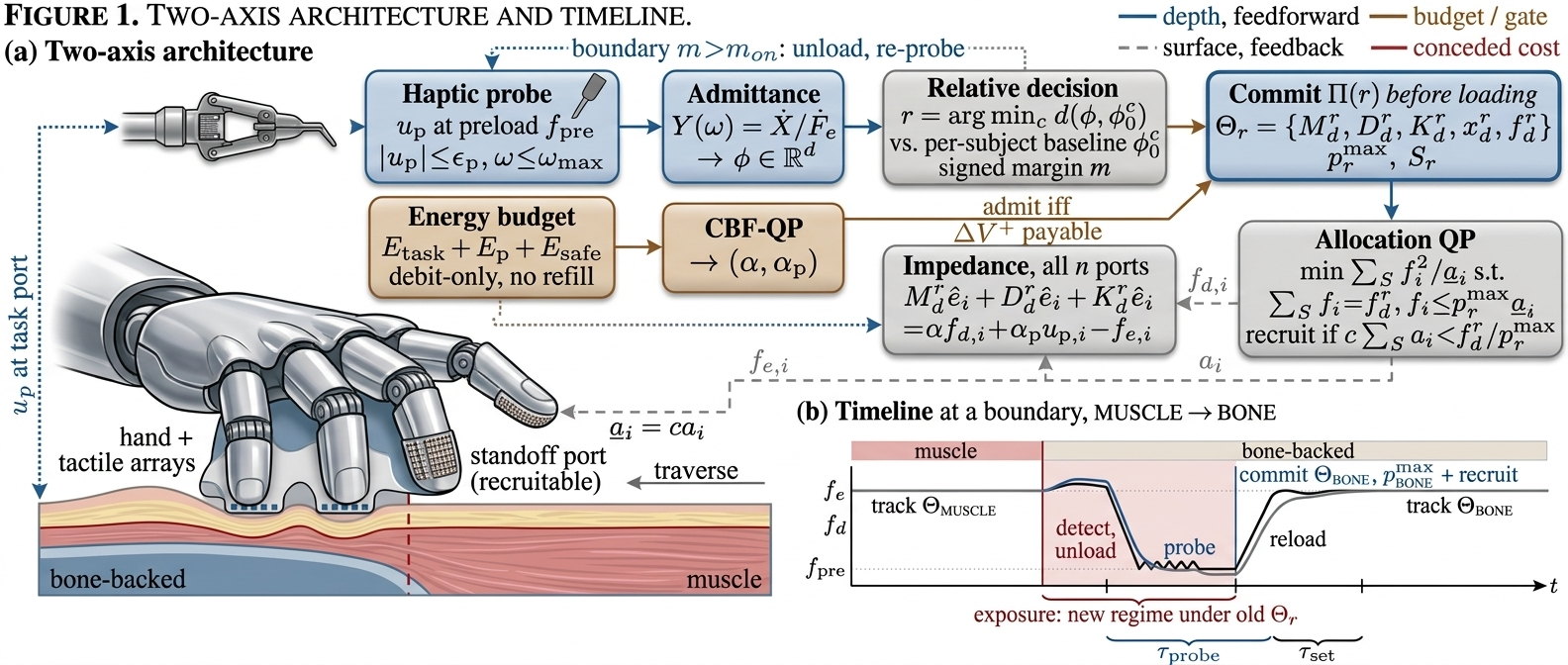}
\caption{(a) Two-axis architecture: the depth axis is feedforward---probe,
discriminate against a per-subject baseline, commit $\Theta_r$ before the load it
governs---while the surface axis tracks the migrating patch in feedback under
that regime. The probe $u_{\mathrm{p}}$ acts at the same port as the task force,
so it appears in~\eqref{eq:imp} and not only in the accounting. The budget is
partitioned three ways, does not refill, and reaches $E_{\mathrm{safe}}$ only by
the conservative switch. (b) Boundary timeline, showing the exposure a reactive
detector leaves.}
\label{fig:framework}
\end{figure}

The allocation of~\eqref{eq:alloc} runs continuously on the measured areas, so it
is feedback on the surface axis, while $\Theta_r$ and $p^{\max}_r$ are
feedforward with respect to each segment's load; the two act on the axes of
Section~\ref{sec:problem} and do not compete. Fig.~\ref{fig:framework} shows the
architecture and the boundary timeline; Algorithm~\ref{alg:main} the per-sample
order of operations.

\begin{algorithm}[htbp]
\caption{Per-sample execution}\label{alg:main}
\begin{algorithmic}[1]
\State \textbf{if} boundary detected ($m>m_{\mathrm{on}}$, Sec.~\ref{sec:probe}) \textbf{then}
\State \quad unload to $f_{\mathrm{pre}}$; probe; classify $r$ \Comment{costs $\tau_{\mathrm{probe}}$}
\State \quad \textbf{if} $\Delta V^{+}$ payable \textbf{then} commit $\Theta_r,p^{\max}_r$ \textbf{else} hold
\State read $a_i$ from tactile arrays; $\underline{a}_i\!\gets\!c\,a_i$
\State \textbf{while} \eqref{eq:areq} infeasible \textbf{and} recruitment admitted \textbf{do}
\State \quad engage next port; debit $\Delta V_j^{+}$ from $E_{\mathrm{task}}$
\State \textbf{if} \eqref{eq:areq} still infeasible \textbf{then}
       $f_d\!\gets\!p^{\max}_r\!\sum_{i\in S}\underline{a}_i$ \Comment{Prop.~\ref{prop:failsafe}}
\State $\{f_{d,i}\}\gets$ solve \eqref{eq:alloc} \Comment{QP, $|S|$ variables}
\State $(\alpha,\alpha_{\mathrm{p}})\gets$ solve safety QP \Comment{Sec.~\ref{sec:passivity}}
\State apply \eqref{eq:imp} at all $n$ ports
\end{algorithmic}
\end{algorithm}

Both programs are small and structurally fixed. Equation~\eqref{eq:alloc} has
$|S|\le n$ variables under one equality and box constraints, and its
Karush--Kuhn--Tucker conditions give the unconstrained minimiser in closed form,
so only the ports whose bound is active need iterating. The safety program is a
feasibility problem in the two scalars $(\alpha,\alpha_{\mathrm{p}})$ over
half-spaces that always contain the origin (Section~\ref{app:proofs}), so it
never fails and admits an active-set solution of the same size. Neither grows
with the trajectory or the number of past switches, which is what makes the
per-sample cost constant rather than merely small.

\subsection{Passivity with a partitioned energy budget}\label{sec:passivity}
The task sources $\alpha f_{d,i}$, the probes $\alpha_{\mathrm{p}}
u_{\mathrm{p},i}$ and the stored-energy jump at a switch all add active power. We
route all three through a budget with a control-barrier-function power
constraint in the manner of~\cite{michel2024energytank} (see
also~\cite{ames2017cbf} and \cite{11127533}), which returns $(\alpha,\alpha_{\mathrm{p}})$ from a
quadratic program. Each sub-budget is its own barrier $h_k(E)=E_k$ and is
\emph{debited only}, $\dot{E}_k=-\max(\beta_k c_k,0)$ with $\beta_k$ the scaling
that drains it---the traverse retracts and the probe oscillates, so a signed
drain would credit the budget back. With a linear class-$\mathcal{K}$ function the
program is then feasible at every state and $\{E_k\ge0\}$ is forward invariant
(Appendix~\ref{app:proofs}).

The layer is otherwise blind to intent: it would meter the authorized probe like
a disturbance and could clamp it. We therefore \emph{partition} the budget,
\begin{equation}
E \;=\; E_{\mathrm{task}} + E_{\mathrm{p}} + E_{\mathrm{safe}},
\qquad E_{\mathrm{task}},E_{\mathrm{p}},E_{\mathrm{safe}}\ge 0,
\label{eq:partition}
\end{equation}
each drained only by its own channel: $E_{\mathrm{task}}$ by the task source and
by switches away from the conservative regime, $E_{\mathrm{p}}$ by probing,
$E_{\mathrm{safe}}$ only by the conservative
$\textsc{muscle}\!\to\!\textsc{bone}$ switch. This is load-bearing: pooled, a
run of switches can starve the next probe and the wrong direction can consume
what the conservative one needed. The budget is finite and \emph{not}
refilled---harvesting $D_d^{\sigma}\dot{x}^2$ would let the designer manufacture
budget by choosing a gain---so $E(0)$ is provisioned in advance, which keeps
$\alpha_{\mathrm{p}}=1$ feasible for every provisioned window
(Appendix~\ref{app:proofs}).

Switching $\Theta_r$ changes $M_d^{r}$, $K_d^{r}$ and $x_d^{r}$ discontinuously. Each
engaged port stores
\begin{equation}
V^i_r=\tfrac{1}{2}M_d^{r}\dot{x}_i^{2}
     +\tfrac{1}{2}K_d^{r}\bigl(x_i-x_{d,i}^{r}\bigr)^{2},
\qquad V_r=\sum_{i=1}^{n}V^i_r ,
\label{eq:V}
\end{equation}
and the port power is $\sum_i f_{e,i}\dot{x}_i$, unengaged ports contributing
nothing to either sum. Each $(x_i,\dot{x}_i)$ is
continuous by \textbf{A3} but the error coordinate is not, $x_{d,i}$ jumping by
$\delta_i=x_{d,i}^{r}-x_{d,i}^{r'}$ while nothing physical moves; evaluating
$V_{r'}-V_{r}$ at the common state (Appendix~\ref{app:proofs}) gives
\begin{equation}
\Delta V=\sum_{i\in S}\Bigl[\tfrac{1}{2}\bigl(M_d^{r'}\!-M_d^{r}\bigr)\dot{x}_i^{2}
+\tfrac{1}{2}\bigl(K_d^{r'}(e_i+\delta_i)^{2}-K_d^{r}e_i^{2}\bigr)\Bigr].
\label{eq:dV}
\end{equation}
The sum runs over $S$ although~\eqref{eq:V} accounts over all $n$ ports: by
\textbf{A5} an unengaged port sits at zero state, so its parameters may change
without repricing any storage. Neither $D_d$ nor $f_d$ appears: $D_d$ enters only
the dissipation rate, and
$f_d$, though it jumps with $\Theta_r$, is metered as a \emph{power}, a finite
jump in which contributes no energy at an instant. A positive $\Delta V$ is
energy the switch creates, so a switch is admitted only when its channel can pay
for it.

\begin{proposition}[Pay-before-switch]\label{prop:pay}
Under \textbf{A3}, \textbf{A5} and \textbf{A6}, let every switch be admitted only if $\Delta V^{+}=\max(\Delta V,0)$ can be paid
from its channel without driving that sub-budget negative, and be debited on
admission. Then $W = V_{\sigma} + E_{\mathrm{task}} + E_{\mathrm{p}} +
E_{\mathrm{safe}}$ is non-increasing at every switch instant, and for every
switching sequence and every provisioned probing schedule the energy delivered
at the interaction port over $[0,T]$ is bounded by $W(0)$.
\end{proposition}

Recruitment is a second class of discrete event, paid on the same terms.
Engaging port $j$ changes only $x_{d,j}$, so with $M_d^{r},K_d^{r}$ fixed inside the
regime the displacement $\delta_j=x_{d,j}^{-}-x_{d,j}^{+}$ gives
\begin{equation}
\Delta V_j \;=\; K_d^{\,r}e_j\delta_j \;+\; \tfrac{1}{2}K_d^{\,r}\delta_j^{2},
\label{eq:dVrec}
\end{equation}
whose cross term is first order in $\delta_j$ and of either sign, so it survives
evaluation at a fixed $e_j$. Releasing a port reverses the displacement and is
free, and reallocating $\{f_{d,i}\}$ costs nothing by the argument that excuses
$f_d$ above.

\begin{proposition}[Refusal is fail-safe]\label{prop:failsafe}
Charge $\textsc{muscle}\!\to\!\textsc{bone}$ to $E_{\mathrm{safe}}$, every other
switch and every recruitment to $E_{\mathrm{task}}$, and on a refused
recruitment reduce the resultant target to $f_d\leftarrow p^{\max}_r\sum_{i\in
S}\underline{a}_i$. Then \eqref{eq:areq} holds whatever the budget state, since
the fallback changes only $f_d$ and so needs no payment; and if, over a traverse
of path length $L$, $E_{\mathrm{safe}}(0)\ge\lceil L/\ell_{\min}\rceil
(\overline{\Delta V}^{\,m\to b}+n_{\max}\overline{\Delta V}_{\mathrm{rec}})$,
then the conservative switch is never refused. Any refusal therefore leaves the
controller in the force-capping regime and within the pressure bound, at the cost
of conformance and force tracking respectively.
\end{proposition}

\noindent The conservative switch tightens $p^{\max}_r$ and so generally requires
recruitment; the pair is admitted as one transaction, whence the port count
$n_{\max}$. The other factor bounds how often the boundary can be crossed: the
method re-commits at most once per $\ell_{\min}$ of path
(Appendix~\ref{app:proofs}), so a traverse of length $L$ admits at most
$\lceil L/\ell_{\min}\rceil$ commitments. We therefore assume the path is known,
not the crossings. Both event classes draw on the same unrefilled
$E_{\mathrm{task}}$, and patch migration makes recruitment the more frequent.

\noindent Three caveats. The bound assumes nothing about the environment, so an
active muscle contraction does not break it but voids its reading as a safety
statement while it lasts. The $\{V_r\}$ are not a common storage function, which
is why payment is needed; augmenting the state with the
budget~\cite{liberzon2003switching} over a dissipative parameter
switch~\cite{zotovic2005impact} and the passivity of~\cite{ott2014hybrid} we do
not claim as new. And a finite budget licenses that much injection rather than
guaranteeing little. Proofs are in Appendix~\ref{app:proofs}.

\section{Experiment}\label{sec:experiments}\label{sec:experiment}


\subsection{Experiment Setup}\label{sec:setup}
\begin{figure}[htbp]
    \centering
    \begin{subfigure}[b]{0.20\columnwidth}
        \centering
        \includegraphics[width=\textwidth, height=3cm]{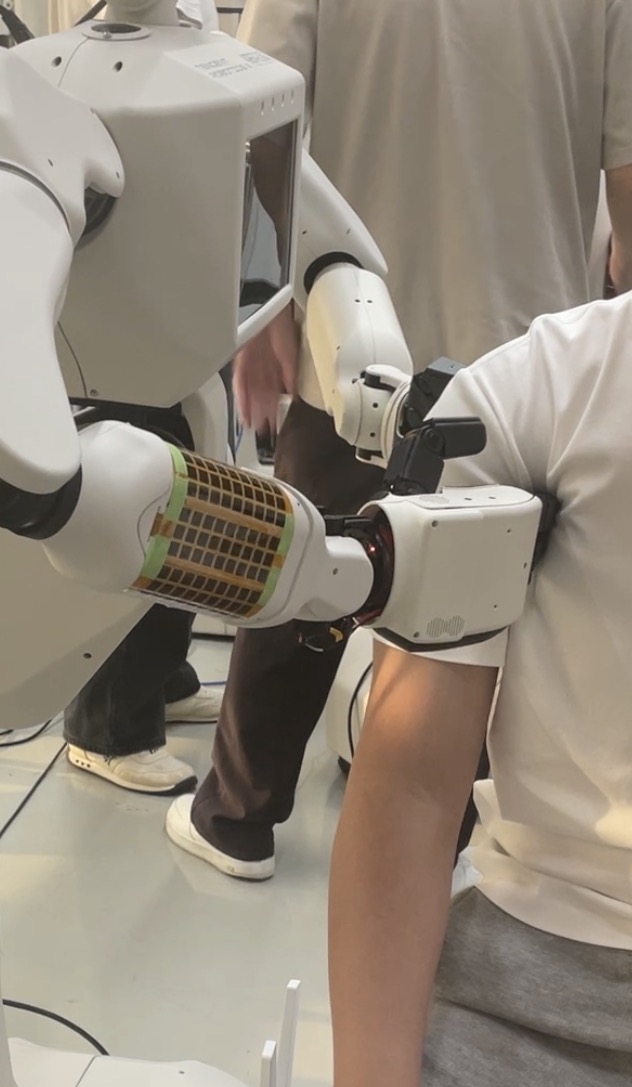}
        \caption{Anterior touch}
        \label{fig:front}
    \end{subfigure}
    \hfill
    \begin{subfigure}[b]{0.20\columnwidth}
        \centering
        \includegraphics[width=\textwidth, height=3cm]{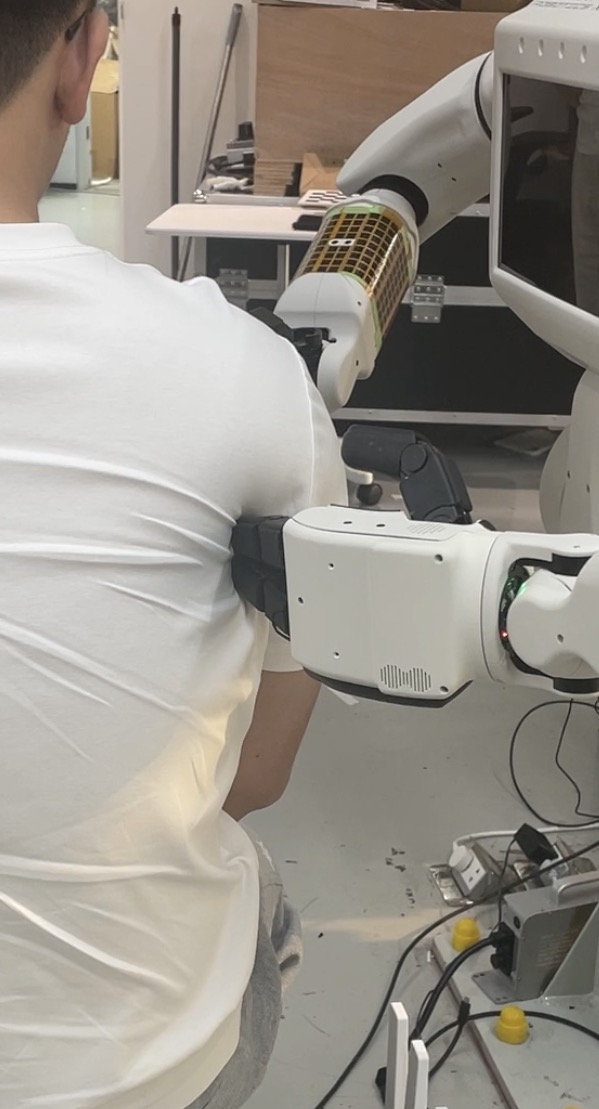}
        \caption{Poster. touch}
        \label{fig:back}
    \end{subfigure}
    \hfill
    \begin{subfigure}[b]{0.20\columnwidth}
        \centering
        \includegraphics[width=\textwidth, height=3cm]{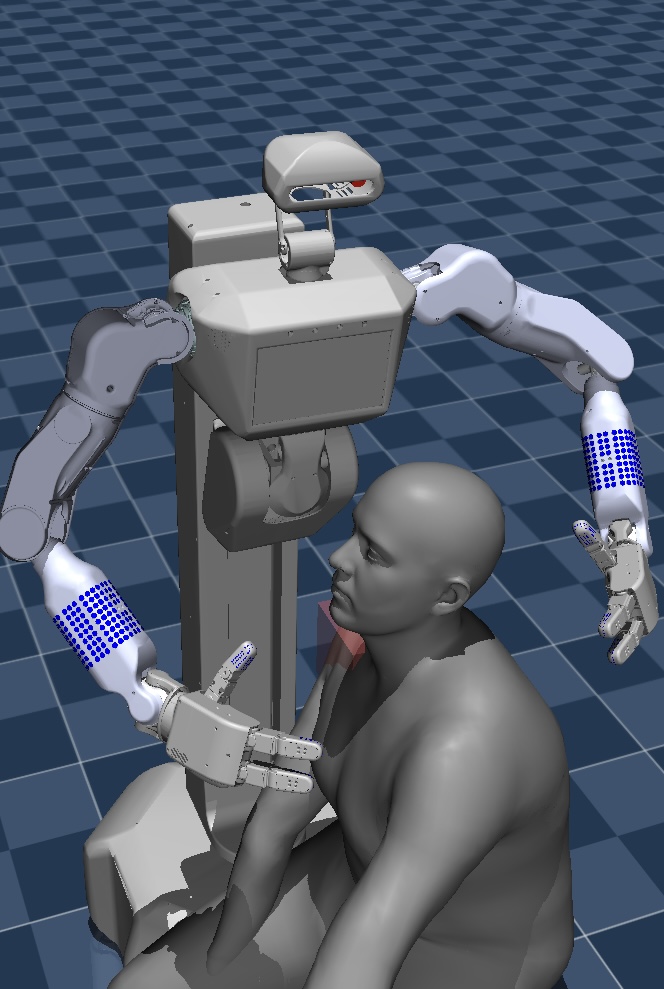}
        \caption{Simulation}
        \label{fig:sim}
    \end{subfigure}
    \hfill
    \begin{subfigure}[b]{0.28\columnwidth}
        \centering
        \includegraphics[width=\textwidth, height=3cm]{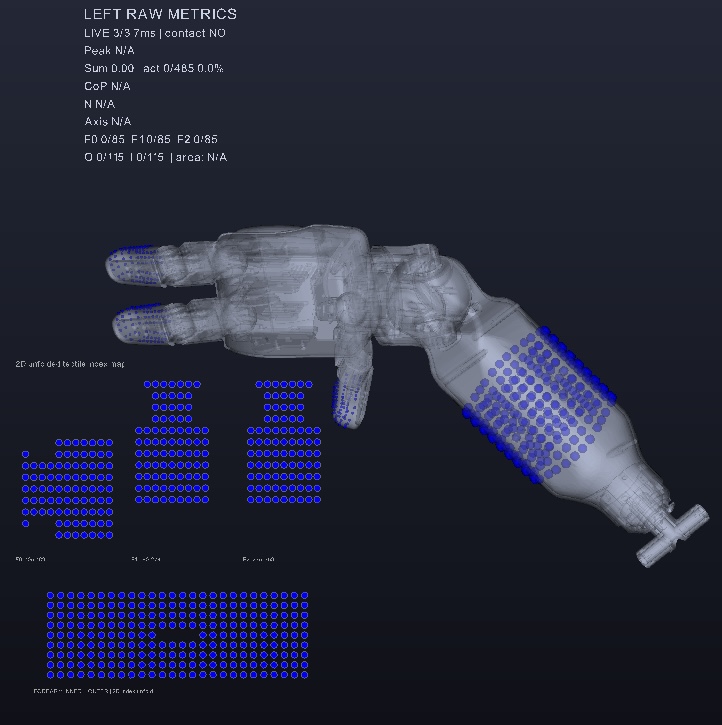}
        \caption{Tactile array}
        \label{fig:tactile}
    \end{subfigure}
    
    \caption{Experimental setup overview. (a, b) Dual-arm physical human experiments showing anterior and posterior contact. (c) Corresponding simulation environment. (d) Schematic illustration of the tactile sensor arrays distributed on the fingers and forearm of the robotic system.}
    \label{fig:experiment_setup}
\end{figure}

In our experiments, a dual-arm system equipped with three-fingered hands
performs fingertip pressing and kneading on the shoulders of human subjects, as
shown in Fig.~\ref{fig:experiment_setup}. A fingertip patch is small, so a modest
resultant already produces high pressure. A small patch is
also sensitive to area loss. Because
the kneading motion keeps the patch migrating, maintaining contact area is what
holds the pressure down. Each fingertip carries a tactile array of resolution $9.593\,\mathrm{mm^2}$,
from which $a_i$ and $f_{e,i}$---hence $p_i$ and $U$---are read; the
allocation~\eqref{eq:alloc}, the feasibility test~\eqref{eq:areq} and
the safety program all run at $50\,\mathrm{Hz}$ per hand.

We compare all three modes on each of three different individuals:

\begin{itemize}
\item \textbf{Conditioned (ours).} The full method: probe before task
loading, commit $\Theta_r$ and $p^{\max}_r$ per segment, allocate
by~\eqref{eq:alloc}, recruit to satisfy~\eqref{eq:areq}, and price every
switch and recruitment against the partitioned budget.
\item \textbf{Baseline.} A single fixed compliance and a single
admissible pressure $p^{\max}=140\,\mathrm{kPa}$ held for the whole
traverse, with no probe, no recruitment and no accounting---the
one-regime controller of Section~\ref{sec:related}.
\item \textbf{Agnostic.} Reactive adaptation from the loading signal
alone, with an intermediate fixed bound $p^{\max}=94\,\mathrm{kPa}$ and
recruitment enabled but no feedforward commitment. This isolates the
contribution of \emph{when} the regime is resolved from that of
recruitment itself.
\end{itemize}

\begin{table}[htbp]
\caption{Force-matched performance, pooled over traverses per
mode. 
}
\label{tab:results}
\centering
\footnotesize
\setlength{\tabcolsep}{1.9pt}
\begin{tabular}{llccccc}
\toprule
& & $|f_e\!-\!f_d^{\,r}|$ & $\max_i p_i$ & $U$ & $A$ & $\max_i p_i$ \\
& Mode & [N] & [kPa] & [--] & [mm$^2$] & $>\!48$\,kPa \\
\midrule
\multirow{3}{*}{\rotatebox[origin=c]{90}{\footnotesize slack}}
 & Baseline    & 4.42\,\tiny{[4.18,\,4.73]} & 34.9\,\tiny{[31.5,\,37.4]} & 1.376\,\tiny{[1.29,\,1.47]} & 77 & 0.15\,$\pm$\,0.17 \\
 & Agnostic    & 3.52\,\tiny{[3.04,\,4.04]} & 31.0\,\tiny{[27.9,\,34.7]} & 1.365\,\tiny{[1.29,\,1.44]} & 58 & 0.12\,$\pm$\,0.15 \\
 & Conditioned & \textbf{2.75}\,\tiny{[2.20,\,4.05]} & 30.9\,\tiny{[29.1,\,34.1]} & 1.355\,\tiny{[1.27,\,1.43]} & 67 & 0.10\,$\pm$\,0.15 \\
\midrule
\multirow{3}{*}{\rotatebox[origin=c]{90}{\footnotesize binding}}
 & Baseline    & 1.66\,\tiny{[1.05,\,2.15]} & 50.2\,\tiny{[47.4,\,62.3]} & 1.252\,\tiny{[1.14,\,1.65]} & 134 & 0.68\,$\pm$\,0.22 \\
 & Agnostic    & 2.19\,\tiny{[0.75,\,2.25]} & 49.1\,\tiny{[47.6,\,228]}  & 1.380\,\tiny{[1.28,\,1.83]} & 125 & 0.68\,$\pm$\,0.28 \\
 & Conditioned & \textbf{1.32}\,\tiny{[1.13,\,1.65]} & \textbf{45.1}\,\tiny{[42.2,\,54.0]} & \textbf{1.161}\,\tiny{[1.11,\,1.22]} & 125 & \textbf{0.45}\,$\pm$\,0.40 \\
\bottomrule
\end{tabular}
\end{table}

\subsection{Three-mode comparison}\label{sec:compare}
\begin{figure}[htbp]
\centering
\includegraphics[width=\columnwidth]{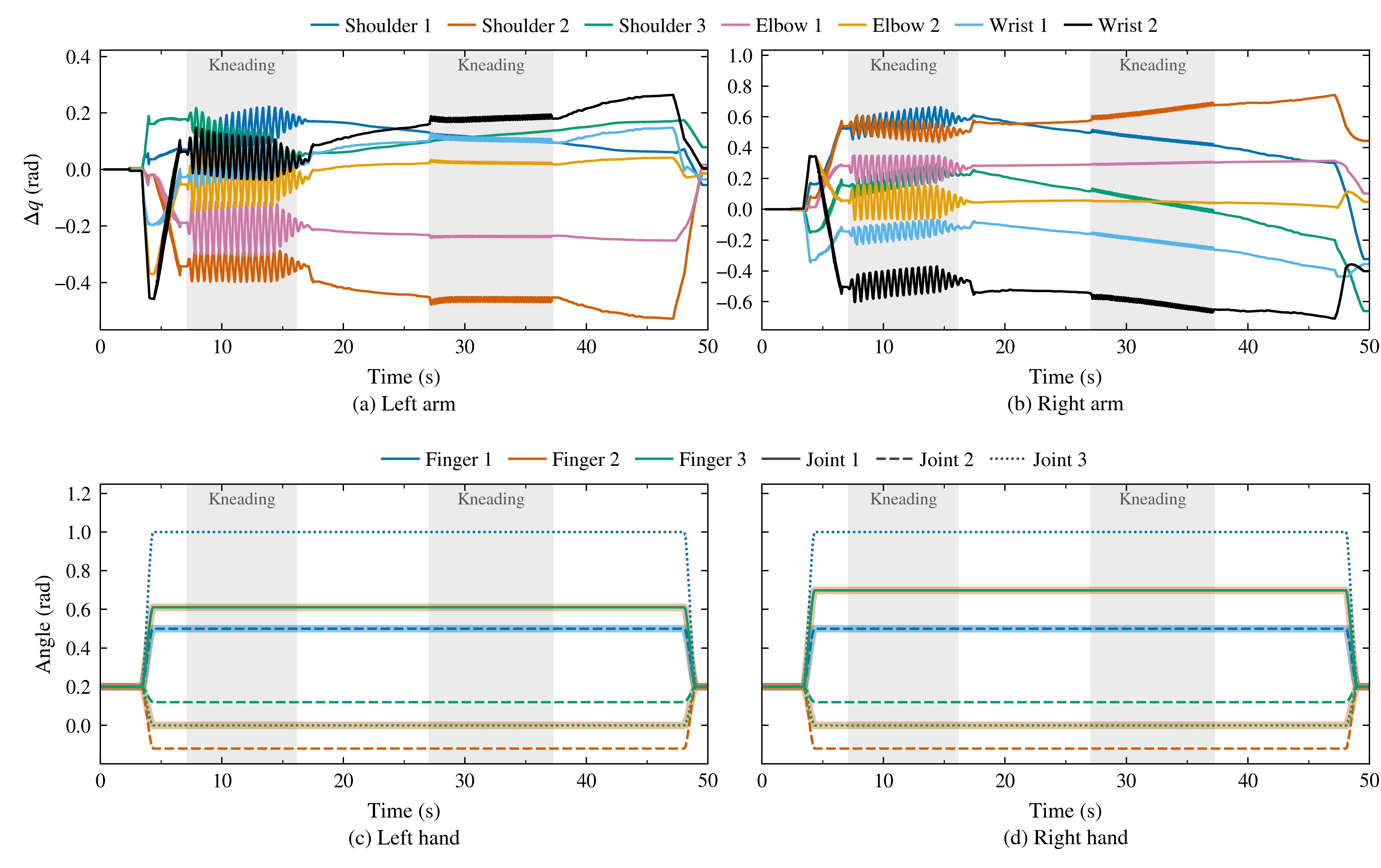}
\caption{Commanded joint trajectories of the pressing-and-kneading motion,
shared by all three modes. (a)~Left and (b)~right arm, shown as displacement
$\Delta q$ from the initial configuration. (c)~Left and (d)~right hand; color
denotes the finger and line style the joint from base to tip, and wide
translucent lines mark joints whose commands coincide with another joint
(Finger~1 joints~1--2; Fingers~2 and~3 at joints~1 and~3). Shaded regions mark
the two kneading phases (7.1--16.2~s, $\approx$2~Hz; 27.0--37.3~s,
$\approx$5~Hz); the finger commands hold a constant grasp throughout.}
\label{fig:jointtraj}
\end{figure}

\begin{figure}[htbp]
\centering
\includegraphics[width=\columnwidth]{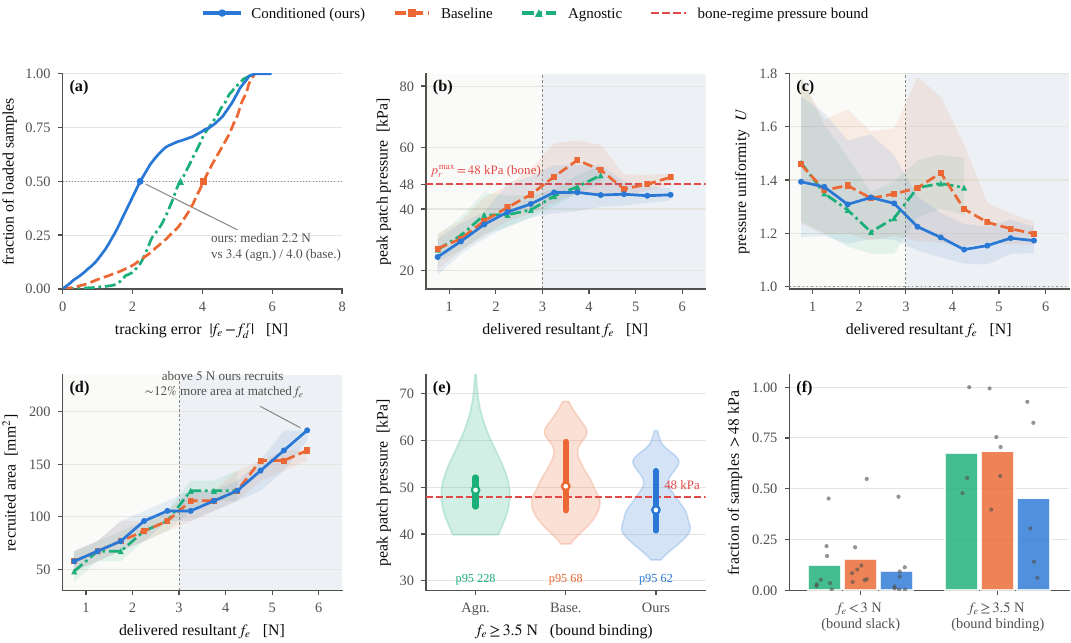}
\caption{All three control modes over the full dataset: eight hand-traverses per mode. \textbf{(a)} Empirical CDF of the
tracking error against the \emph{committed} resultant $f_d^{\,r}$. \textbf{(b)}--\textbf{(d)} Force-matched profiles---binned
median (markers) and interquartile band against the delivered resultant
$f_e$, drawn only where a bin holds $\ge150$ samples, which is why the
agnostic mode (whose traverses rarely reach high resultants) stops near
$4.3\,\mathrm{N}$. \textbf{(e)} Peak-pressure distribution
where the bound binds; violin bodies span the $2$--$95$th percentile
range with each mode's $95$th percentile printed beneath. Bars mark the interquartile range, open circles the median.
\textbf{(f)} Fraction of loaded samples whose peak pressure exceeds the
bone bound, by load band; dots are individual hand-traverses. Results are split at $f_e = 3\,\mathrm{N}$ (untuned), where $\max_i p_i$ reaches $48\,\mathrm{kPa}$, separating slack ($<3\,\mathrm{N}$) and binding ($>3\,\mathrm{N}$) bone bounds.}
\label{fig:compare}
\end{figure}

Fig.~\ref{fig:compare} and Table~\ref{tab:results} give the pooled
comparison. All three modes execute the same commanded joint trajectory for the
pressing-and-kneading motion (Fig.~\ref{fig:jointtraj}), so the differences
reported below arise from the controller rather than the motion.

\textbf{Force Tracking.} Against the resultant it commits to, the conditioned
mode realizes a median error of $2.22\,\mathrm{N}$ over all loaded
samples, with a $95\%$ interval of $[1.86,2.83]$, against
$3.37\,\mathrm{N}$ $[2.96,3.88]$ for the agnostic mode and
$4.04\,\mathrm{N}$ $[3.59,4.42]$ for the baseline---a $45\%$ reduction on
the baseline. Per hand-traverse the
mean absolute errors are $2.63\pm0.83$, $3.27\pm0.53$ and
$3.86\pm0.70\,\mathrm{N}$. The
interpretation is the one Section~\ref{sec:control} argues for: because
$\Theta_r$ is committed before the segment is loaded, the controller
starts each segment from the right parameters instead of converging to
them, and the error it does not accumulate during that convergence is
what the margin consists of.

\textbf{Force Distribution.} Below the slack bone bound, the three modes coincide (medians within $0.021$), validating the high-load result by reflecting an uninformative low-load regime rather than a mere offset. Where the bound binds, the
conditioned mode separates. Uniformity falls to $1.161$ with an interval
of $[1.11,1.22]$, disjoint from the agnostic mode's $[1.28,1.83]$;
against the baseline the point estimate is $7\%$ lower.
Peak pressure behaves the same way: $45.1\,\mathrm{kPa}$ against $50.2$
and $49.1$, and Fig.~\ref{fig:compare}(b) shows the mechanism---the
conditioned median never crosses $48\,\mathrm{kPa}$ anywhere in the
binding band, while the baseline crosses near $3.2\,\mathrm{N}$ and the
agnostic mode near $4.2\,\mathrm{N}$. The tails matter more than the
medians here, and Fig.~\ref{fig:compare}(e) is where they show: the
$95$th percentile of peak pressure is $62\,\mathrm{kPa}$ for our method,
$68$ for the baseline and $228$ for the agnostic mode. Fig.~\ref{fig:compare}(d) shows the
area side of the same trade: above $5\,\mathrm{N}$ the conditioned mode
holds $182\,\mathrm{mm^2}$ at matched resultant against the baseline's
$163$, roughly $12\%$ more, which is how the lower pressure at equal
force is produced. In exceedance terms (Fig.~\ref{fig:compare}(f))
bone-unsafe pressure is reached on $45\%$ of loaded high-force samples
under our method against $68\%$ for both alternatives; restricted to the
hand whose tactile array does not clip the
same figures are $6\%$, $70\%$ and $52\%$, and the recruited area there
is $173\,\mathrm{mm^2}$ against the baseline's $144$.

\section{Conclusion}\label{sec:conclusion}
We have shown that the subsurface regime of layered soft tissue can be resolved
by haptic probing before task loading and consumed feedforward by a
force-distribution law. The probe is read against a per-subject baseline. The
committed regime sets the admissible pressure and, through~\eqref{eq:areq}, the
area the hand must recruit. 

Against a single-regime baseline and a reactive controller, the median error
against the committed resultant was $2.22\,\mathrm{N}$, versus $4.04$ and
$3.37\,\mathrm{N}$. This is a $45\%$ reduction on the baseline, with disjoint
$95\%$ intervals. Where the bone bound binds, uniformity was $1.161$ against
$1.252$ and $1.380$. The $95$th-percentile peak pressure was $62\,\mathrm{kPa}$
against $68$ and $228\,\mathrm{kPa}$, achieved by recruiting about $12\%$ more
area at matched resultant.

Boundary detection remains reactive, so each new regime is briefly loaded under
the previous parameters. The unrefilled budget also caps the recruitments per
traverse. Look-ahead probing, graded regimes and peak-taxel regulation are the
natural next steps.

\section{Proofs and bounds}\label{app:proofs}

Throughout, $x_i$ increases into the tissue and $f_{e,i}\ge0$ opposes
penetration, so port power is $\sum_i f_{e,i}\dot{x}_i$. Write
$\sigma(t)\in\{\textsc{bone},\textsc{muscle}\}$ for the regime signal, $V_r$ as
in~\eqref{eq:V}, and $W=V_{\sigma}+E_{\mathrm{task}}+E_{\mathrm{p}}
+E_{\mathrm{safe}}$. Sums run over all $n$ ports, unengaged ones contributing
zero. The sub-budgets are debited only,
$\dot{E}_{\mathrm{task}}=-\max(\alpha c_{\mathrm{task}},0)$ and
$\dot{E}_{\mathrm{p}}=-\max(\alpha_{\mathrm{p}}c_{\mathrm{p}},0)$ with
$c_{\mathrm{task}}=\sum_i f_{d,i}\dot{x}_i$ and
$c_{\mathrm{p}}=\sum_i u_{\mathrm{p},i}\dot{x}_i$, and
$\dot{E}_{\mathrm{safe}}=0$, plus the discrete debits of the admission rule.

\emph{Jump formula~\eqref{eq:dV}.} Position and velocity are continuous at a
switch, since the switch changes controller parameters rather than applying an
impulse. The error coordinate is not: with $e_c=x-x_d^{c}$, eliminating the
common $x(t_k)$ gives $e_{r'}=e_r+\delta$, $\delta=x_d^{r}-x_d^{r'}$, while
$\dot{e}$ is unchanged. Evaluating $V_{r'}-V_{r}$ at the common physical state
yields~\eqref{eq:dV}. Holding $e$ fixed instead would drop the cross-term
$K_d^{r'}e\,\delta$, which is first order in $\delta$ and of either sign.

\emph{Within a regime.} Differentiating $V^i_r$ along~\eqref{eq:imp} gives
$f_{e,i}\dot{x}_i=-\dot{V}^i_r+\alpha f_{d,i}\dot{x}_i
+\alpha_{\mathrm{p}}u_{\mathrm{p},i}\dot{x}_i-D_d^{r}\dot{x}_i^2$. Summing over
$i$, each active term is at most the drain of its own sub-budget---$\alpha
c_{\mathrm{task}}\le\max(\alpha c_{\mathrm{task}},0)=-\dot{E}_{\mathrm{task}}$,
and likewise for the probe, the inequality being strict exactly when a retraction
would otherwise credit the budget---so
$\sum_i f_{e,i}\dot{x}_i\le-\dot{W}-\sum_i D_d^{r}\dot{x}_i^2\le-\dot{W}$.
Truncating therefore costs only conservatism. The bound holds whether or
not probing is active: this is why the probe must appear in~\eqref{eq:imp}, since
an accounting that meters a signal absent from the dynamics would leave the
probing windows---the case the method exists to protect---outside the proof.

\emph{Across a switch.} The physical state and the unaffected sub-budgets are
continuous, so $\Delta W=\Delta V-\Delta V^{+}=\min(\Delta V,0)\le0$ under the
admission rule; a refused switch leaves $W$ unchanged. Telescoping the
within-regime inequality over the switching instants and using $W(T)\ge0$ gives
$\int_0^T\sum_i f_{e,i}\dot{x}_i\,\mathrm{d}t\le W(0)$ for every switching sequence, however
dense, and every provisioned probing schedule. No assumption on the environment
is used: the bound constrains the controller side of the port. An active muscle
contraction therefore does not break it---the tissue becomes a second source,
which makes the left side smaller while voiding the bound's reading as a safety
statement, since $W(0)$ no longer caps the energy circulating at the port.

\emph{Recruitment.} Engaging port $j$ leaves $(x_j,\dot{x}_j)$ continuous
(\textbf{A3}) and changes only $x_{d,j}$, so with $M_d^{r},K_d^{r}$ fixed inside the
regime $\Delta V_j=V^j_r(e_j+\delta_j)-V^j_r(e_j)$ yields~\eqref{eq:dVrec}, the
remaining ports being untouched. Admission debits $\Delta V_j^{+}$ exactly as a
switch does, so $\Delta W\le0$ and the telescoping argument is unchanged. On
refusal, setting $f_d=p^{\max}_r\sum_{i\in S}\underline{a}_i$
makes~\eqref{eq:areq} hold with equality, so~\eqref{eq:alloc} is feasible; and
since only $f_d$ changes, no storage is repriced and the reduction itself needs
no payment. The pressure bound therefore holds at every budget state, the
resultant absorbing the cost.

\emph{Barrier feasibility.} Each $\dot{E}_k=-\max(\beta_k c_k,0)$ is piecewise
affine in its scaling $\beta_k\in\{\alpha,\alpha_{\mathrm{p}}\}$, with $c_k$
independent of it because \eqref{eq:alloc} allocates the nominal target. Where
$c_k\le0$ the barrier condition $\dot{h}_k\ge-\lambda h_k$ holds for every
$\beta_k$; where $c_k>0$ it reads $\beta_k c_k\le\lambda E_k$, a half-space of
relative degree one. Both contain $\beta_k=0$, so the program is feasible at
every state. The gain $\lambda$ sets how sharply a sub-budget is throttled as it
empties: the condition binds once $E_k\le\beta_k c_k/\lambda$, so $\lambda$ is
chosen so that this threshold sits below the smallest transaction the traverse
must still afford, $\lambda\ge\max_k c_k^{\max}/\underline{E}$ with
$\underline{E}$ that reserve. A larger $\lambda$ defers throttling at the cost of
approaching $E_k=0$ more closely; the invariance conclusion holds for any
$\lambda>0$.

\emph{Probe authorization.} With $|u_{\mathrm{p}}|\le\varepsilon_{\mathrm{p}}$
band-limited to $\omega_{\max}$ and $\bar{G}$ bounding the closed-loop gain from
probe force to normal velocity over that band,
$|P_{\mathrm{probe}}|\le\varepsilon_{\mathrm{p}}^{2}\bar{G}$, so a window costs at
most $\varepsilon_{\mathrm{p}}^{2}\bar{G}T_{\mathrm{win}}$. Amplitude alone would
not suffice, the velocity response growing with frequency. As $E_{\mathrm{p}}$ is
drained by nothing else, provisioning it for the planned window count keeps
$\alpha_{\mathrm{p}}=1$ feasible throughout.

\emph{Fail-safe refusal.} $E_{\mathrm{safe}}$ is drained by no other channel, and
the cadence bound below caps a traverse of length $L$ at
$\lceil L/\ell_{\min}\rceil$ commitments, hence at that many conservative
switches; each costs at most $\overline{\Delta V}^{\,m\to b}$ plus $n_{\max}$
recruitments, so provisioning $E_{\mathrm{safe}}(0)$ for that many keeps the
reserve solvent for the whole traverse and only the non-conservative direction can be refused. A refused
recruitment triggers the $f_d$ reduction above, which needs no payment, so the
pressure bound never depends on solvency. Holding \textsc{bone} over muscle costs
conformance, not safety.

\emph{Bounds on cadence.} Passivity requires no dwell time; two other
requirements do. Hysteresis: proposing a switch only when~\eqref{eq:margin}
exceeds $m_{\mathrm{on}}>\sigma_m$, the margin measurement-error bound, leaves a
locus with $m\le0$ proposing nothing, so label noise cannot chatter at a
boundary. Transient completion: $\tau_{\mathrm{set}}=\max_r
4/|\mathrm{Re}\,\lambda^r_{\mathrm{slow}}|$, with $\lambda^r_{\mathrm{slow}}$ the
slower root of $M_d^r\lambda^2+D_d^r\lambda+K_d^r=0$; the familiar $8M_d^r/D_d^r$
holds only when underdamped and would underestimate settling in the heavily
damped regime our own tissue argument implies. With $\tau_{\mathrm{probe}}$ the
full unload--probe--decide--reload cycle,
$\tau_d\ge\max(\tau_{\mathrm{set}},\tau_{\mathrm{probe}})$ and the method
re-commits at most once per $\ell_{\min}\le v_{\max}\tau_d$ of path. Because the
budget does not refill, admitted transactions---switches and
recruitments---are capped at $\lfloor E_{\mathrm{task}}(0)/\overline{\Delta
V}\rfloor$, $\overline{\Delta V}$ being the envelope bound on $\Delta V^{+}$. That
bound is a design-time constant, not an online quantity: on the operating box
$|\dot{x}_i|\le v_{\max}$, $|e_i|\le e_{\max}$, dropping the non-positive
$-\tfrac{1}{2}K_d^{r}e_i^{2}$ from~\eqref{eq:dV} and maximising the rest gives
\begin{equation}
\begin{split}
\overline{\Delta V}_{(r\to r')}=\tfrac{n}{2}\Bigl[&\max\bigl(0,M_d^{r'}\!-M_d^{r}\bigr)v_{\max}^{2}\\[-2pt]
&+K_d^{r'}\bigl(e_{\max}+|\delta|_{\max}\bigr)^{2}\Bigr],
\end{split}
\label{eq:dVbar}
\end{equation}
and $\overline{\Delta V}=\max_{(r,r')}\overline{\Delta V}_{(r\to r')}$. With
$|\mathcal{R}|=2$ there are two ordered pairs, so this is the larger of two
numbers fixed once the two parameter sets are chosen; $\overline{\Delta
V}_{\mathrm{rec}}$ follows the same way from~\eqref{eq:dVrec} with $n=1$.


{\footnotesize
\bibliographystyle{IEEEtran}
\bibliography{refs}}

\end{document}